\documentclass[letterpaper, 10 pt, conference]{ieeeconf}  

\IEEEoverridecommandlockouts                              

\usepackage{lipsum}
\usepackage{hyperref}
\usepackage[nolist]{acronym}
\usepackage{siunitx}

\usepackage{amsmath}
\usepackage{amssymb}
\usepackage{mathabx}
\usepackage{graphicx}
\usepackage[caption=false,font=footnotesize]{subfig}

\usepackage{booktabs}
\usepackage{multirow}

\usepackage[usenames,dvipsnames]{xcolor}
\usepackage[most]{tcolorbox}
\usepackage[T1]{fontenc}
\usepackage{import}
\usepackage[latte]{catppuccin}

\usepackage{cite} 

\usepackage{mdframed}
\usepackage{soul}
\tcbuselibrary{skins}

\pagecolor{white}
\color{black}
\tcbset{%
    colback=ctpBase,
    boxrule=0mm,
    left=1mm,
    top=1mm,
    right=1mm,
    bottom=1mm,
    boxsep=0mm,
    before skip=0mm,
    after skip=0mm,
    arc=3mm,
    opacityframe=0,
    interior hidden
}

\makeatletter
\newtcolorbox{picturebox}[2][]{%
enhanced,frame hidden,interior hidden,fonttitle=\bfseries,
overlay={\begin{tcbclipframe}\node at (frame)
{\includegraphics{#2}};\end{tcbclipframe}%
},#1}
\makeatother

\begin{acronym}
    \acro{RMSE}{root mean squared error}
    \acro{GBP}{Gaussian Belief Propagation}
    \acro{GNN}{Graph Neural Network}
    \acro{NMPC}{Nonlinear Model-Predictive Control}
    \acro{WT}{waypoint tracking}
    \acro{PT}{path tracking}
    \acro{SP}{structured planner}
    \acro{PPD}{perpendicular path deviation}
    \acro{ORCA}{Optimal Reciprocal Collision Avoidance}
    \acro{SDF}{Signed Distance Field}
\end{acronym}

\title{\LARGE \bf Multi-Agent Path Planning in Complex Environments using Gaussian Belief Propagation with Global Path Finding
}

\author{Jens Høigaard Jensen$^{1}$, Kristoffer Plagborg Bak Sørensen$^{1}$, Jonas le Fevre Sejersen$^{1}$, and Andriy Sarabakha$^{1}$%
\thanks{$^{1}$J. Høigaard Jensen, K. Plagborg Bak Sørensen, J. le Fevre and A. Sarabakha are with the Department of Electrical and Computer Engineering, Aarhus University, 8000 Aarhus C, Denmark, {\tt\small jens.jens@live.com, kristoffer.pbs@gmail.com, \{jonas.le.fevre, andriy\}@ece.au.dk.}}%
}

\begin{document}

\maketitle
\thispagestyle{empty}
\pagestyle{empty}

\begin{abstract}
Multi-agent path planning is a critical challenge in robotics, requiring agents to navigate complex environments while avoiding collisions and optimizing travel efficiency. This work addresses the limitations of existing approaches by combining Gaussian belief propagation with path integration and introducing a novel tracking factor to ensure strict adherence to global paths. The proposed method is tested with two different global path-planning approaches: rapidly exploring random trees and a structured planner, which leverages predefined lane structures to improve coordination. A simulation environment was developed to validate the proposed method across diverse scenarios, each posing unique challenges in navigation and communication. Simulation results demonstrate that the tracking factor reduces path deviation by 28\% in single-agent and 16\% in multi-agent scenarios, highlighting its effectiveness in improving multi-agent coordination, especially when combined with structured global planning.

\end{abstract}

\section{INTRODUCTION}

The evolution of automation has played a fundamental role in advancing both industry and society since the Industrial Revolution. As technology continues to progress, autonomous systems are reaching new levels of complexity and functionality. These advancements are particularly evident in sectors such as logistics, warehouse management, and autonomous transportation, where multi-agent systems are increasingly employed to improve operational efficiency. For instance, autonomous vehicles demonstrate the potential to enhance traffic flow through coordination and communication between vehicles. In constrained environments, such as warehouses or bounded road networks, multi-agent systems must operate efficiently, avoiding deadlocks and collisions. Enhancing the capabilities of these systems will enable future technologies to perform more effectively, conserving both time and resources. 

Effective multi-agent path planning is central to the success of these systems, which optimizes the routes taken by individual agents to achieve their objectives while avoiding collisions. This necessitates sophisticated approaches to coordination, communication, and collision avoidance, ensuring that agents can navigate their environments without interfering with each other. 

Current systems often encounter difficulties in collision avoidance and efficient navigation in such settings. In particular, the field of \ac{GBP} collaborative planning lacks the ability to ensure strict adherence to the global path, which is crucial for optimizing multi-agent coordination over longer routes. While global path planning provides a high-level framework for efficient navigation, maintaining adherence to these paths in dynamic, multi-agent environments introduces significant challenges.

This work addresses the limitations of existing multi-agent path planning systems, particularly in complex and constrained environments. A novel approach that integrates a tracking factor is proposed, to enhance the overall efficiency of multi-agent navigation. The tracking factor ensures precise adherence to pre-determined global paths, allowing the system to optimize navigation efficiency and collision avoidance in real time.
\begin{figure}[t]
    \centering
    \includegraphics[width=0.98\linewidth]{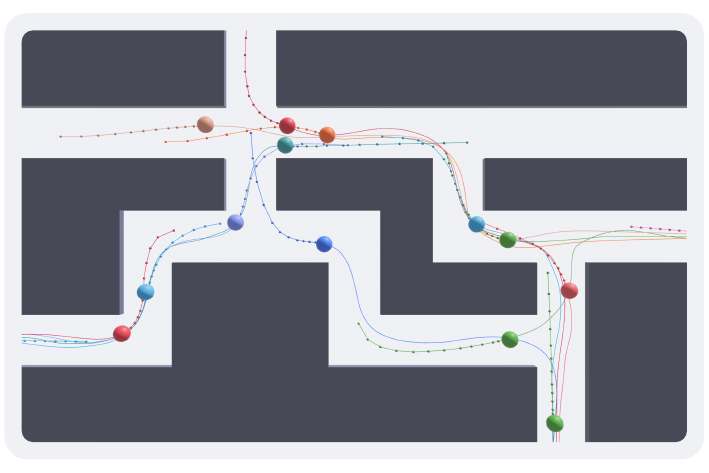}
    \caption{Illustration of the Complex environment with multiple crossings and bends. Larger spheres represent agents, while the chains denote variables in agents' factor graphs and the solid lines indicate agents' traveled path.}
    \label{fig:abstract}
\end{figure}
\medbreak
\noindent The key contributions of this study are as follows: 
\begin{enumerate}
    \item Global path planning has been integrated into the \ac{GBP} Planner~\cite{GBPPlanner}, enhancing multi-agent efficiency and coordination in complex environments. 
    \item A novel tracking factor is introduced to ensure precise path adherence, improving navigation accuracy and reducing deviations in multi-agent planning. 
    \item A multi-agent \ac{GBP} simulation framework\footnote{All source code for this work is made publicly available on \href{https://github.com/au-Master-Thesis/magics}{GitHub} under the MIT license.} has been developed to validate and refine the proposed \ac{GBP} algorithms across diverse scenarios, offering a foundation for further research and optimization in multi-agent systems.
\end{enumerate}
\newpage
\section{RELATED WORKS}
Traditionally, path planning has been approached using centralized and distributed methodologies, each with strengths and limitations. Centralized methods, in which a single controller coordinates all robots’ paths, have shown high performance in structured environments. These approaches often rely on global information and can efficiently handle large-scale coordination. However, as the number of robots grows or as environments become more dynamic and complex, centralized methods face scalability issues due to their computational demands. For example, methods like priority-based planning \cite{soria2021predictive, okumura2022priority, machines10090773, le2023cameta} have demonstrated strong performance but require detailed, global knowledge of the environment and agent locations, which limits their applicability in more dynamic and unpredictable settings.

To address the limitations of centralized planning, distributed methods have gained significant attention. One widely-used distributed multi-robot planner is the \ac{ORCA} algorithm \cite{NH-OCRA, snape2010smooth, claes2012collision}, which allows robots to modify their velocities based on the positions of neighboring agents. While \ac{ORCA} has been successful in many real-world applications, it often produces inefficient and jerky trajectories in dense environments due to its reliance on instantaneous velocity adjustments without long-term planning. Other distributed methods, like \cite{alonso2018cooperative}, rely on short-term trajectory predictions but fail to account for the entire look-ahead time window, making them less effective in highly dynamic environments.


More recent works have explored the use of graph-based methods, which enable more flexible communication between agents. Approaches utilizing the alternating direction method of multipliers (ADMM) \cite{van2016online} and distributed model predictive control (MPC) \cite{luis2020online} have shown the ability to maintain long-term trajectory optimization through iterative communication. However, these methods often rely on ideal communication conditions and can struggle with intermittent or unreliable networks. Building on the ADMM framework, a distributed trajectory optimization approach has been proposed to address asynchronous communication and message delays \cite{ferranti2022distributed, tordesillas2020mader}. This method adjusts safety margins and leverages predictive strategies to ensure collision-free trajectories, even in the face of packet loss. Although it enhances robustness to communication faults, the approach introduces conservatism, potentially leading to less efficient paths and longer completion times.

In recent years, learning-based methods have also been explored for multi-robot coordination. \Ac{GNN} \cite{li2020graph, niu2021multi} have been applied to optimize local communication and decision-making, particularly in grid-world environments. While \ac{GNN}s can handle local information sharing efficiently, their application to more complex environments has been limited. Additionally, methods such as reinforcement learning-based systems \cite{s23125615, 10368056, paul2023efficient} have explored the optimization of task allocation and path planning in multi-agent systems, but they often face challenges in scalability and robustness, particularly when communication failures occur.

Another promising approach is the integration of differentiable decentralized planners like D2CoPlan \cite{sharma2023d2coplan}, which allows for the efficient management of multi-robot coverage by incorporating both local and global objectives. Similarly, scalable systems that leverage large language models for multi-robot collaboration \cite{chen2024scalable} have begun to explore how decentralized systems can balance between efficiency and scalability. However, these methods still remain in early development stages.

A promising development in the field of multi-agent path planning is the use of \ac{GBP} for multi-agent path planning. \ac{GBP} allows for distributed computation over factor graphs, where each robot can communicate and optimize its path with local neighbors through message passing. The Robot Web project \cite{RobotWeb} first demonstrated the potential of \ac{GBP} for localization in large multi-robot systems, showing that robots could iteratively share and update their beliefs about positions. This concept was later extended with a \ac{GBP}-based planner \cite{GBPPlanner}, a purely distributed technique formulated using a generic factor graph to handle both dynamics and collision constraints over a forward time window. However, while a \ac{GBP}-based planner provides a strong framework for decentralized communication, like many other distributed methods, it focuses primarily on local collision avoidance and does not inherently incorporate global path planning, which can lead to suboptimal navigation over longer trajectories.

To address these limitations, our work integrates global path planning into the \ac{GBP} framework. Our approach introduces a tracking factor that ensures robots adhere to global paths, improving both navigation robustness towards communication failures and collision avoidance in multi-agent systems. This method is particularly effective in complex environments, where the integration of global and local planning components is essential for scalable, real-world deployments. 
\section{METHODOLOGY}

\ac{GBP} relies on factor graphs to model the multi-agent path planning problem, where the states of robots are represented as variables, and the constraints or dependencies between these states are captured as factors \cite{dellaert_factor_2021}. These factors impose soft constraints on one or more variables, guiding the agents' movements to satisfy all imposed constraints as effectively as possible. The key mechanism within \ac{GBP} is message passing; a probabilistic inference technique where each node, representing a variable or factor, iteratively exchanges information with neighboring nodes. These messages encode the node's beliefs about the variables' values, which are updated based on the information received. Over successive iterations, this process converges to a joint distribution that optimally satisfies the graph's constraints, determining the most likely future position and velocity in the plane for each robot \cite{ortiz_visual_2021}.

The factors introduced in the \ac{GBP} planner include four essential factors: (i) the pose factor $f_p$ represents the robot's estimated position and orientation; (ii) the dynamics factor $f_d$, captures the relationship between consecutive poses, ensuring consistency with the robot's motion model; (iii) the obstacle factor $f_o$, imposes constraints to avoid collisions based on pixel sampling from a signed distance field image representation of the environment; (iv) and the inter-robot factor $f_i$, which accounts for the relative positioning and interaction between multiple robots in a collaborative setting. These factors are parameterized by the scalar value $d_i$ which specifies the distance, from which the factor is active.

The original algorithm in \cite{GBPPlanner} works well on a local level but lacks a global overview of how to get from $A$ to $B$, which results in many local optima\footnote[1]{Video comparison of the three methods: \href{https://www.youtube.com/watch?v=Uzz57A4Tk5E}{video}}. To solve this, a global pathfinding algorithm has to be leveraged. The proposed method is algorithm-agnostic; any path-finding algorithm that outputs a series of waypoints avoiding obstacles can be used. 
This global plan can be introduced into the factor graph using two different approaches.

\subsection{Approach 1: waypoint tracking}
The \ac{WT} approach serves as the baseline in this work, leveraging a global planner to generate a sequence of waypoints that act as intermediate navigation goals for the local planning algorithm. This approach automates the process of global path planning, which was previously done manually in \cite{GBPPlanner}. \ac{WT} remains a foundational method for integrating global pathfinding into multi-agent planning, using the same core variables and factors from the original \ac{GBP} planner, but now with automated waypoint generation that dynamically adapts to the environment. The process consists of the following steps: 
\begin{enumerate} 
    \item A global planner computes a path from the start position $A$ to the goal position $B$, accounting for environmental constraints and obstacles. 
    \item The path is segmented into waypoints, which act as intermediate targets for the local planner.
    \item The existing local planning algorithm follows these waypoints without any modification to its core operation. 
\end{enumerate}
The approach is designed to guide the robot towards the waypoints along the given path, but it does not strictly enforce adherence to the exact obstacle-free line between waypoints. This flexibility allows the robots to deviate from the globally planned path, enabling more creative maneuvering around each other and permitting corner-cutting. While this can provide some advantages in dynamic environments, it may also result in an increase in intersecting paths among robots, particularly in densely populated areas. This makes it an effective baseline for evaluating more accurate approaches, such as path tracking, which prioritizes stricter adherence to global paths.


\subsection{Approach 2: path tracking}
\begin{figure}[!b]
    \centering
    \vspace{-4mm}
    \subfloat[The tracking factor pulls the variable towards and along the path, with a green area near waypoint $w_1$ indicating corner tracking.]{%
        \includegraphics[width=0.65\linewidth]{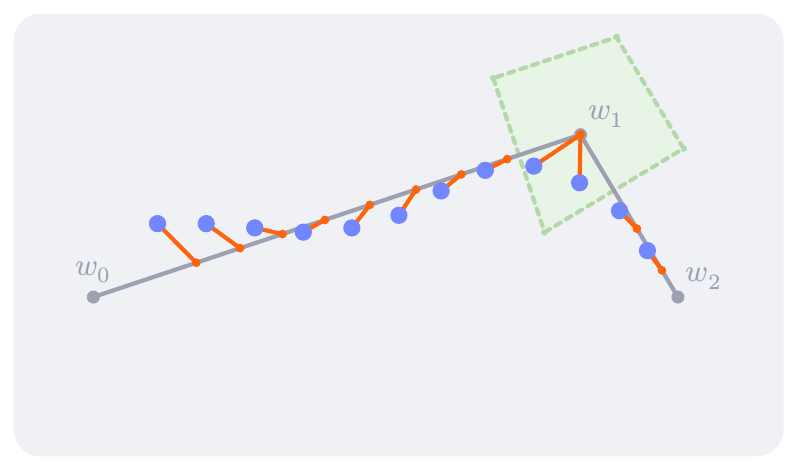}%
        \label{fig:tracking-factor-explain-1}%
    }
    \hfill
    \subfloat[Tracking factors for robot R moving from $w_0$ to $w_1$. ]{%
        \includegraphics[width=0.33\linewidth]{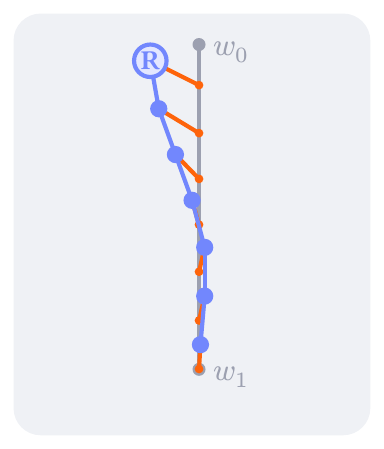}%
        \label{fig:tracking-factor-explain-2}%
    }
    \caption{Illustration of the tracking factor concept. The tracking factor measurement is shown in orange, the state variables is shown in blue, and the gray line is the global path connecting two waypoints.}
    \label{fig:tracking-design}
\end{figure}

The \ac{PT} approach builds on the foundation of global planning by introducing a tracking factor, $f_t$, to enforce stricter adherence to the planned path. This tracking factor is integrated into the factor graph structure used in \ac{GBP}, ensuring that robots follow the global path more closely. The $f_t$ attaches to each variable within the robot's prediction horizon, with the exception of the first and last variables, which are anchored to ensure stability at the starting and goal positions.

Fig. \ref{fig:tracking-design} demonstrates how the tracking factor operates. It measures the perpendicular distance from the robot's current position to the planned path and applies a pulling force that not only reduces the deviation from the path but also nudges the robot slightly forward, particularly around corners. This forward pull helps to guide the robot smoothly along turns, ensuring more precise navigation when compared to the more flexible \ac{WT} approach.

\subsection{Tracking factor}
The tracking factor $f_t$ exerts a pulling force that guides the variable toward the desired path, ensuring adherence to the planned path. 
This is achieved through two key components: the measurement function and the Jacobian, which guide the robot along the global path while respecting its dynamics and other constraints.

\subsubsection{Measurement Function}
The measurement function evaluates the deviation of the variable's current position, \( \mathbf{x}_\mathrm{pos} = [x\ y]^\top \), from a desired path $\textbf{GP} = \{\textbf{p}_1, \dots, \textbf{p}_n\}$, where $\textbf{p}_i = [x_i, y_i]$. Each line segment along the path is defined as \( \mathbf{l}_i = \mathbf{p}_{i+1} - \mathbf{p}_i \), where the index \( i \) identifies the current segment the system is tracking. The output of the measurement function is a scalar value between 0 and 1, which dictates the strength of the pulling effect that guides the variable along the trajectory.

The core concept of the tracking factor is to project the current position of the variable \( \mathbf{x}_\mathrm{pos} \) onto the line segment \( \mathbf{l}_i \). The distance between this projection and the actual position serves as the basis for the measurement. The projection function computes the projection of the point \( \mathbf{x} \) onto the line segment between \( \mathbf{p}_i\) and \( \mathbf{p}_{i+1} \), and is defined as follows:
\begin{equation}
    \small
\mathcal{P}_i = \mathbf{p}_i + \frac{(\mathbf{x} - \mathbf{p}_i) \cdot (\mathbf{p}_{i+1} - \mathbf{p}_i)}{\|\mathbf{p}_{i+1} - \mathbf{p}_i\|^2} (\mathbf{p}_{i+1} - \mathbf{p}_i).
\end{equation}
As the variable approaches the waypoint $\textbf{p}_{i+1}$, the transition to the next line segment occurs when the variable enters a predefined radius around the waypoint, controlled by the configurable threshold \( r_{\text{switch}} \). At this point, the index \( i \) is incremented, i.e., \( i = i+1 \).

Once \( i \) is updated, a conditional criterion, denoted by \( q \), determines whether the variable is transitioning between two line segments:
\begin{equation}
    q = \|\mathcal{P}_i - \mathbf{p}_i\| < r_{\text{switch}} \land \|\mathcal{P}_{i-1} - \mathbf{p}_i\| < r_{\text{switch}}\label{eq.condition},
\end{equation}
\noindent where \( \mathcal{P}_i \) is the projection onto the current line segment, and \( \mathcal{P}_{i-1} \) is the projection onto the previous segment.
Based on this condition, the measurement point \( \mathbf{x}_\mathrm{meas} \) is defined as:
\begin{equation}
    \small
    \mathbf{x}_\mathrm{meas} =
    \begin{cases}
        \mathbf{x_\mathrm{pos}} + \frac{1}{2}((\mathcal{P}_i - \mathbf{x}_\mathrm{pos}) - (\mathcal{P}_{i-1} - \mathbf{x}_\mathrm{pos}))\, &\mathrm{if}\ q \\
        \mathcal{P}_i + \mathbf{d} \cdot \frac{||\mathbf{x}_\mathrm{vel}||}{s_v}\, &\mathrm{else}
    \end{cases} \label{eq.meas-point},
\end{equation}
\noindent where \( \mathbf{d} = \frac{\mathbf{l}_i}{\|\mathbf{l}_i\|} \) is the normalized direction vector of the current line segment, and the term \( \mathbf{d} \cdot \frac{\|\mathbf{x}_\mathrm{vel}\|}{s_v} \) ensures that the system maintains movement along the segment, not just perpendicular correction. The parameter \( s_v\), chosen heuristically, balances forward momentum and prevents overshooting. The condition \( q \) serves two key purposes. First, it facilitates a smooth transition between the path segments \( \textbf{p}_{i-1}\) and \( \textbf{p}_i \), enabling the system to consider both segments in subsequent iterations. Second, it pulls the system toward the corner formed by the intersection of the two segments, effectively reducing corner-cutting behavior. This is illustrated in Fig.~\ref{fig:tracking-factor-explain-1}.

Finally, the measurement function \( h(\mathbf{x}) \), over the state vector \( \mathbf{x} \in \mathbb{R}^{4\times 1} = [\mathbf{x}_{\text{pos}}, \mathbf{x}_{\text{vel}}]^\top \), computes a scaled and clamped distance between the current position \( \mathbf{x}_\mathrm{pos} \) and the measurement point \( \mathbf{x}_\mathrm{meas} \):
\begin{equation}
   h(\mathbf{x}) = \min\left(1, \frac{\|\mathbf{x}_{\text{pos}} - \mathbf{x}_{\text{meas}}\|}{d_a}\right)
\end{equation}
The raw distance is normalized by the parameter \( d_a \), which governs how quickly the attraction force reaches its maximum. The result is clamped to a maximum of 1 to ensure stability in the factor graph inference and prevent excessively large attraction forces.

\subsubsection{Jacobian}
The Jacobian is used to compute the next position of the variable, imposed by the constraint represented by the tracking factor. The most recent measurement, \( h \), scales the Euclidean difference between \( \mathbf{x}_\mathrm{meas} \) and \( \mathbf{x}_\mathrm{pos} \), ensuring the velocity of the system is accounted for. The Jacobian is defined as follows:
\begin{equation}
    \mathbf{J} = \left[ \frac{1}{h(\mathbf{x})}(x_\mathrm{meas} - x_\mathrm{pos})\ \frac{1}{h(\mathbf{x})}(y_\mathrm{meas} - y_\mathrm{pos})\ 0\ 0 \right] \label{eq.jacobian}.
\end{equation}
The Jacobian is padded with zeros to incorporate the velocity component of the linearization point.

\begin{figure}
    \centering
    \includegraphics[width=0.7\linewidth]{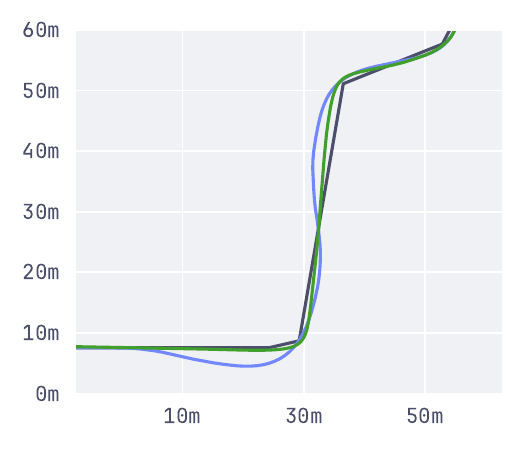}
    \vspace{-3mm}
    \caption{The path deviation visualized between waypoint tracking (blue), path tracking (green), and desired path (black).}
    \label{fig:tracking}
\end{figure}
By leveraging this tracking factor, the \ac{PT} approach should significantly reduce deviations from the planned trajectory, resulting in more accurate path-following behavior. This is especially beneficial in scenarios where high precision is critical, such as in densely populated environments or when robots must closely coordinate to avoid collisions.
While the structured nature of \ac{PT} enhances multi-agent coordination and is ideal for environments demanding precise path tracking and collision avoidance, it does impose stricter trajectory-following constraints. This can limit the robot's ability to make creative maneuvers or adjust to unforeseen obstacles in real-time making it less adaptable in highly dynamic settings where frequent adjustments are required. 
\

\section{EXPERIMENTS}
The performance of the system is assessed based on key metrics that capture aspects such as navigation efficiency, safety, and path accuracy. These experiments are conducted in different environments, each presenting unique challenges to test the robustness of the approaches.

\subsection{Metric}
Several metrics are used to measure efficiency, safety, and accuracy in evaluating our navigation strategies. The following metrics provide a comprehensive assessment of the system's performance.
\begin{itemize}
    \item \textit{Inter Robot Collisions:} Number of collisions between robots. The physical size of each robot is represented by a bounding circle. A collision between two robots happens when their circles intersect. That is a collision is only registered if robots \(R_a\) and \(R_b\) intersect at timestep \(t_n\) but not at \(t_{n-1}\). 
    
    \item \textit{Environment Robot Collisions:}  Number of collisions between robots and the environment. Similar to \textit{Inter Robot Collisions}, bounding circles are used for the robots, while the environment obstacle is equipped with a collider of the same geometric layout.
    
    \item \textit{Root mean squared error of \ac{PPD}:} At each sampled position, the distance between it and the closest projection onto each line segment of the planned path is measured and accumulated, using the \ac{RMSE}, defined as follows:
    \begin{equation}
        \text{RMSE} = \sqrt{\frac{1}{n} \sum_{j=1}^{n} \left( \min_{i} \left\lbrace \|P_j - p(P_j, L_i)\|^2 \right\rbrace \right)^2},
    \end{equation}
    where \(L_i \in \{L_1, L_2, ..., L_m\}\) is the set of line segments making up the planned path, \(P_j \in \{P_1, P_2, ..., P_n\}\) is the set of sampled positions, and \(\| P_j - p(P_j, L_i) \|^2\) is the squared distance between the sampled position \(P_j\) and the projection of \(P_j\) onto the line segment \(L_i\). This is measured to test the effect of the proposed tracking factor, as some applications require that robots follow a dictated path with little deviation. 
\end{itemize}
\subsection{Scenarios}
\label{sec:scenarios}
We introduce three scenarios to test the proposed strategies, each presenting unique challenges in navigation and communication, thereby assessing the robots' robustness in varied conditions. If not specified otherwise, the following parameters are used for each scenario: Robot radius $r_R = 2$m, communication radius $r_\text{comms} = 20$m, communication failure probability $\gamma = 0\%$, target speed $v_t = 5$m/s, time horizon $t_{K-1} = 5$s, number of internal iterations per timestep $T_I = 10$, and external $T_E = 10$. The standard deviation for each factor is $\sigma_{f_d} = 0.1$, $\sigma_{f_p} = 10^{-15}$, $\sigma_{f_i} = 0.005$, $\sigma_{f_o} = 0.005$ and $\sigma_{f_t} = 0.15$. 

\begin{figure}
    \centering
    \subfloat[Junction environment]{%
    
        \begin{tcolorbox}[width=0.4\linewidth]
            \includegraphics[width=\textwidth]{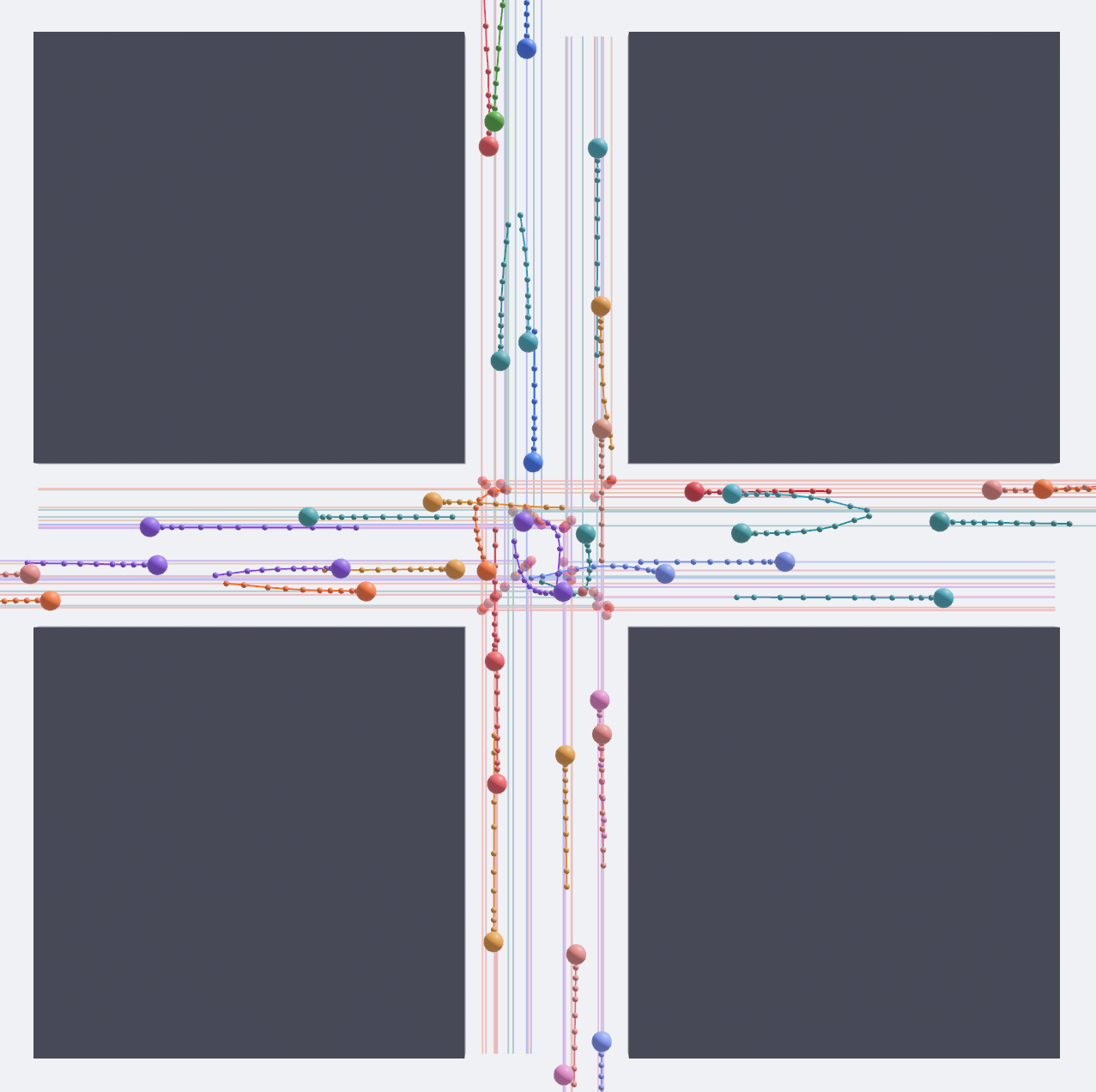}
        \end{tcolorbox}%
        \label{fig:junction}%
    }
    \hfill
    \subfloat[Complex environment]{%
        \begin{tcolorbox}[
            width=0.55\linewidth,
            boxrule=0pt,
            boxsep=2.5pt
        ]
            \includegraphics[width=\textwidth]{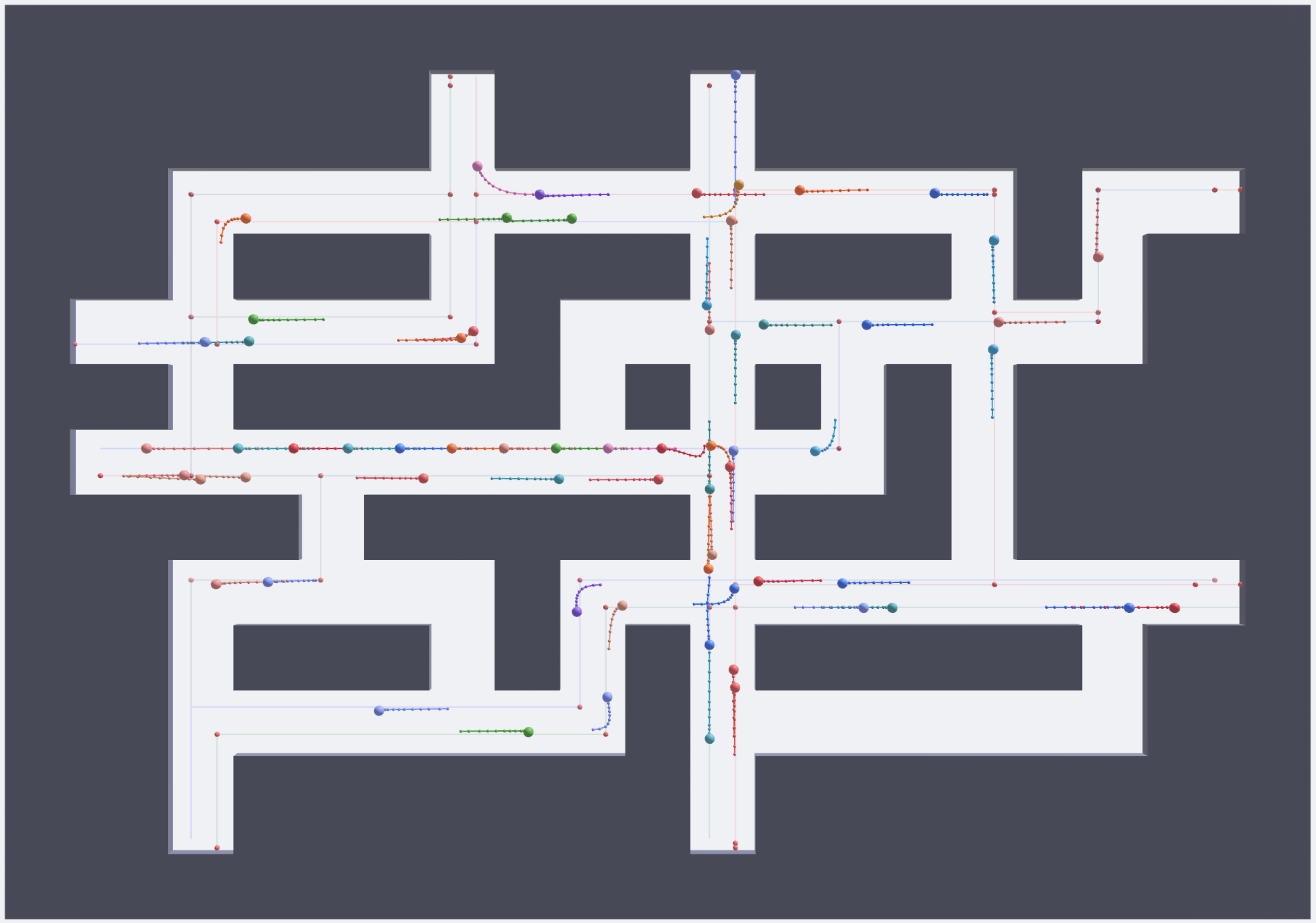}
        \end{tcolorbox}%
        \label{fig:complex_env}%
    }
    \caption{Illustration of the two environments used in the experiments: (a) junction environment, (b) complex environment.}
    \label{fig:tracking-factor-explain-3}
\vspace{-3mm}
\end{figure}

\textbf{The junction environment} consists of a single two-way junction designed to evaluate robot coordination in high-throughput conditions. Each robot is randomly assigned a spawning point on one of the four sides of the junction, with a target destination chosen from the remaining three sides. Consequently, all robots must navigate through the junction, requiring effective coordination to avoid collisions and ensure smooth passage, as shown in Fig. \ref{fig:junction}. To increase robot density in the center, each robot is assigned a smaller radius of $r_R = 1m$.


\textbf{Communications failure} is tested within the junction environment and tests the robustness of each presented strategy in terms of inter robot collisions, environment robot collisions, and RMSE of \ac{PPD} by selecting the degree of communication failure $\gamma$ from $[0\%, 10\%, \dots, 70\%]$. A communication failure is defined as shutting down all communication for a single unit at a given timestep $t_n$. 

\textbf{The complex environment} is a building designed to have a maze-like structure with hallways, junctions and lane merging, creating a challenging navigation environment. What looks like dead ends in Fig. \ref{fig:complex_env} are spawning and goal locations. Each individual robot gets a task to traverse the complex environment to one of the goal locations. 

Two global planners are tested in this environment. The first is the asymptotically optimal RRT* path planning algorithm \cite{karaman2011sampling}, which is employed to automatically generate global paths. However, the inherent randomness of the RRT* algorithm can result in unstructured paths, increasing the likelihood of robot intersections as they navigate the environment. The second global planner is a \ac{SP} that incorporates lane structures to guide the robots along more organized and predictable paths. The environment needs to be predefined as a directed graph, where edges represent lanes going in one direction. This approach is designed to reduce intersections by encouraging robots to follow predefined lanes, thus promoting smoother and more efficient navigation through the environment. To find a valid path through the environment, the A* algorithm can be used to determine the shortest route. For both the solo and collaborative modalities, these parameters are different from the default; $v_t =7$ m/s, $\sigma_{f_i} = \sigma_{f_o} = 10^{-3}$.







\section{RESULTS}
The results presented in Table \ref{table:tracking} for the solo modality offer an isolated assessment of the impact of the tracking factor. First, a single robot navigates through the complex environment using the RRT* global planner. A notable reduction in \ac{PPD} is observed at approximately 28\%. This outcome highlights the potential of the tracking factor when evaluated in isolation.
\begin{table}[b]
    \centering
    \caption{Navigation results in the complex environment.}
    \begin{tabular}{ccc|cc|cc}
        \toprule
        \multirow{2}{*}{\textbf{Modality}} & \multirow{2}{*}{\textbf{Method}} & \multirow{2}{*}{\textbf{Planner}} & \multicolumn{2}{c|}{\textbf{PPD [m]}} & \multicolumn{2}{c}{\textbf{Collisions}} \\
        & & & \textbf{Mean} & \textbf{Std.} & \textbf{Int.} & \textbf{Ext.} \\
        \midrule
        \multirow{2}{*}{\textbf{Solo}}
        & WT & RRT* & 1.03 & 0.11 & 0 & 0 \\
        & PT & RRT* & 0.74 & 0.08 & 0 & 0 \\
        \midrule
        \multirow{4}{*}{\textbf{Collaborative}}
        & WT & RRT* & 1.00 & 0.23 & 2.8 & 0 \\
        & PT & RRT* & 0.86 & 0.25 & 11.9 & 0 \\
        & WT & SP & 0.64 & 0.11 & 0 & 0 \\
        & PT & SP & 0.54 & 0.10 & 0 & 0 \\
        \bottomrule
    \end{tabular}
    \label{table:tracking}
\end{table}

The results for the collaborative modality in the complex environment are also summarized in Table \ref{table:tracking}. When using the \ac{SP}, both the \ac{WT} and \ac{PT} methods successfully avoid collisions, between robots and with the environment. In contrast, when utilizing the sample-based planner RRT*, as in the solo modality, a significant increase in inter-robot collisions occurs, with a modest \ac{PPD} reduction of only 14\% between \ac{PT} and \ac{WT}. This disparity is likely attributed to the random nature of RRT*, which generates independent paths for each robot without considering the paths of others. As a result, many globally planned paths overlap, leading to chaotic and conflicting trajectories that place excessive demands on the local \ac{GBP} planner, ultimately making collisions unavoidable. In contrast, the structured planner, which assigns dedicated lanes for each direction, effectively mitigates these issues, resulting in zero collisions.

When comparing \ac{WT} to \ac{PT} in the collaborative \ac{SP} scenario, \ac{PT} demonstrates an almost 16\% reduction in \ac{PPD} on average, with a corresponding decrease in variance. These improvements are promising, particularly considering that \ac{PT} outperforms the RRT* approach by 2\% while maintaining zero collisions.

\begin{figure}[b]
    \centering
    \includegraphics[width=1\linewidth]{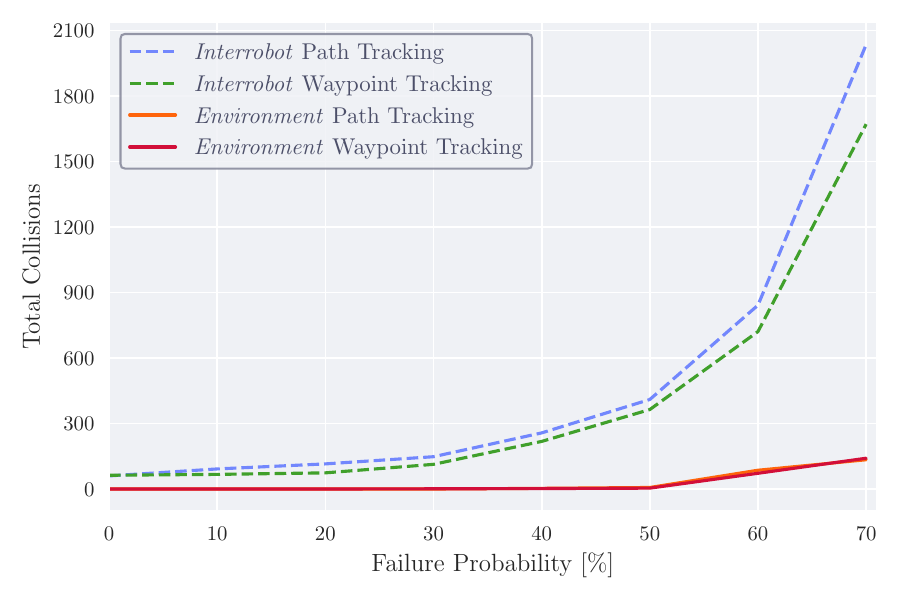}
    \caption{Results from the junction environment. There is a slight increase in inter-robot collisions when using path tracking instead of waypoint. The number of collisions with the environment is the same for WT and PT. Average over 5 runs for each failure rate with 1200 robots.}
    \vspace{-5 pt}
    \label{fig:sjtw-collisions}
\end{figure}

It is important to note that these results are obtained under ideal communication conditions. As illustrated in Fig. \ref{fig:sjtw-collisions}, the tracking factor does influence the number of inter-robot collisions under conditions of increasing communication failure. However, the increase is relatively minor, with significant deviations from \ac{WT} results only occurring when the communication failure rate exceeds 60\%. Thus, the marginal increase in collisions is outweighed by the improvements in path deviation, especially in scenarios where collisions occur regardless. 
An increase in obstacle collision can also be observed when the communication failure rate exceeds 50\%. At this failure rate, few iterations of \ac{GBP} include external information from other robots, which in turn also means that the internal iterations are weighted disproportionately. In this scenario, each robot has optimized for internal costs, which can cause suboptimal external collision avoidance. The weighting between $\sigma_{f_i}$ and $\sigma_{f_o}$ determines whether it is more likely to become an inter robot or environment collision.

\begin{figure}
    \centering
    \includegraphics[width=1\linewidth]{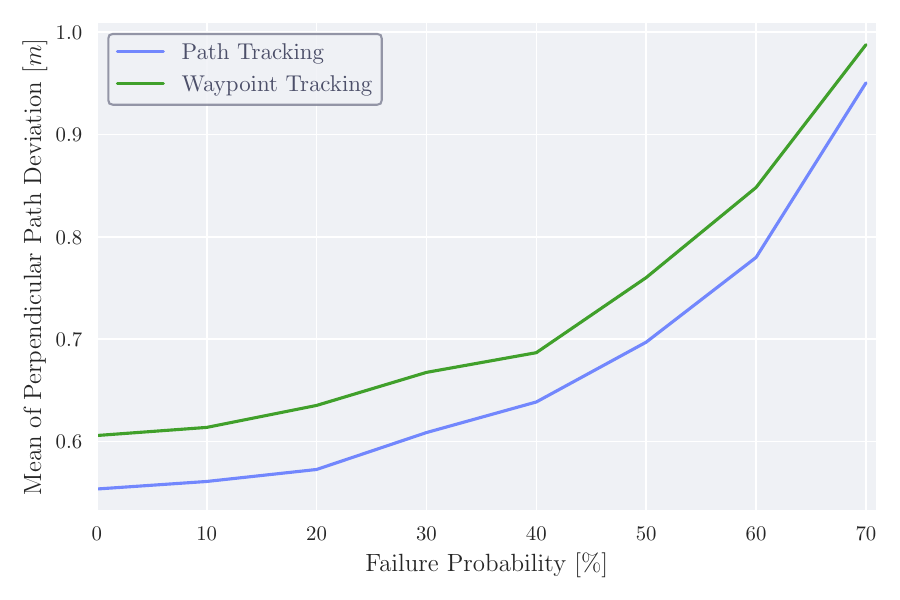}
    \caption{Perpendicular path deviation as a function of the failure probability. Lower is better.}
    \vspace{-5 pt}
    \label{fig:sjtw-ppe}
\end{figure}

The degradation of \ac{PPD} under these conditions, as shown in Fig. \ref{fig:sjtw-ppe}, further illustrates the trade-offs between accuracy and collision avoidance. As the communication failure rate increases, the robots become less coordinated, requiring more agile maneuvers to avoid collisions. This reduces path efficiency, as the robots deviate from optimal trajectories and struggle to maintain smooth, collision-free navigation.


A significant qualitative outcome of this research is, the introduction of the tracking factor $f_t$ offers significant benefits for path adherence, but it requires a delicate tuning process. If the balance between the tracking and dynamics factors is not carefully maintained, there is a risk of the system becoming trapped in local minima. Proper tuning is crucial to prevent the robot from getting stuck at sharp corners or over-correcting due to excessive reliance on the tracking factor. Furthermore, while the tracking factor helps guide the robot along the path, it must work in harmony with the inter-robot factor to avoid collisions. Effective tuning ensures these factors complement each other, optimizing both path adherence and collision avoidance.

The global planning element does not degrade local cooperative collision avoidance, as waypoint tracking automates what was previously manual. However, the random nature of the RRT* algorithm can cause suboptimal crossing paths. Manually placed waypoints can be risk-averse, reducing path crosses. The tracking factor exacerbates this issue by pulling actors towards planned paths without much consideration of others, making collision avoidance difficult. Balancing tracking and inter-robot factors helps, but collisions remain more frequent than with waypoint tracking alone. Solving this is crucial for effective path tracking with tracking factors. Still, we believe in a structured environment with roads and rules, the tracking factor would ensure more predictable and safer navigation. 

\section{CONCLUSIONS}
In this paper, we introduced a novel tracking factor within a \ac{GBP} framework to enhance multi-agent path planning. By incorporating this tracking factor, robots are able to adhere more strictly to global paths, significantly improving navigation efficiency and collision avoidance. Our approach was tested with two global planners: RRT* and \ac{SP}. While the random nature of RRT* resulted in increased inter-robot collisions due to uncoordinated path intersections, the structured planner combined with the tracking factor completely eliminated collisions, highlighting the advantages of path tracking combined with structured global planning. The tracking factor proved especially effective in reducing path deviation and improving multi-agent coordination by ensuring that robots remained aligned with the desired paths even in complex environments. In multi-agent scenarios, this approach demonstrated a 16\% improvement in path deviation, further emphasizing its potential to promote smoother and safer navigation.

\section*{Acknowledgment}
The authors would like to acknowledge the financial and hardware contribution from Beumer Group A/S, and Innovation Fund Denmark (IFD) under File No. 1044-00007B.
\addtolength{\textheight}{-9cm}   

\newpage 
\bibliographystyle{IEEEtran.bst}
\bibliography{References.bib}

\begin{thebibliography}{10}
\providecommand{\url}[1]{#1}
\csname url@rmstyle\endcsname
\providecommand{\newblock}{\relax}
\providecommand{\bibinfo}[2]{#2}
\providecommand\BIBentrySTDinterwordspacing{\spaceskip=0pt\relax}
\providecommand\BIBentryALTinterwordstretchfactor{4}
\providecommand\BIBentryALTinterwordspacing{\spaceskip=\fontdimen2\font plus
\BIBentryALTinterwordstretchfactor\fontdimen3\font minus \fontdimen4\font\relax}
\providecommand\BIBforeignlanguage[2]{{%
\expandafter\ifx\csname l@#1\endcsname\relax
\typeout{** WARNING: IEEEtran.bst: No hyphenation pattern has been}%
\typeout{** loaded for the language `#1'. Using the pattern for}%
\typeout{** the default language instead.}%
\else
\language=\csname l@#1\endcsname
\fi
#2}}

\bibitem{GBPPlanner}
A.~Patwardhan, R.~Murai, and A.~J. Davison, ``Distributing collaborative multi-robot planning with gaussian belief propagation,'' \emph{IEEE Robotics and Automation Letters}, vol.~8, no.~2, pp. 552--559, 2023.

\bibitem{soria2021predictive}
E.~Soria, F.~Schiano, and D.~Floreano, ``Predictive control of aerial swarms in cluttered environments,'' \emph{Nature Machine Intelligence}, vol.~3, no.~6, pp. 545--554, 2021.

\bibitem{okumura2022priority}
K.~Okumura, M.~Machida, X.~D{\'e}fago, and Y.~Tamura, ``Priority inheritance with backtracking for iterative multi-agent path finding,'' \emph{Artificial Intelligence}, vol. 310, p. 103752, 2022.

\bibitem{machines10090773}
S.~Lin, A.~Liu, J.~Wang, and X.~Kong, ``A review of path-planning approaches for multiple mobile robots,'' \emph{Machines}, vol.~10, no.~9, p. 773, 2022.

\bibitem{le2023cameta}
J.~le~Fevre~Sejersen and E.~Kayacan, ``Cameta: Conflict-aware multi-agent estimated time of arrival prediction for mobile robots,'' in \emph{2023 IEEE/RSJ International Conference on Intelligent Robots and Systems (IROS)}.\hskip 1em plus 0.5em minus 0.4em\relax IEEE, 2023, pp. 9254--9261.

\bibitem{NH-OCRA}
J.~Alonso-Mora, A.~Breitenmoser, M.~Rufli, P.~Beardsley, and R.~Siegwart, ``Optimal reciprocal collision avoidance for multiple non-holonomic robots,'' in \emph{Distributed autonomous robotic systems: The 10th international symposium}.\hskip 1em plus 0.5em minus 0.4em\relax Springer, 2013, pp. 203--216.

\bibitem{snape2010smooth}
J.~Snape, J.~Van Den~Berg, S.~J. Guy, and D.~Manocha, ``Smooth and collision-free navigation for multiple robots under differential-drive constraints,'' in \emph{2010 IEEE/RSJ international conference on intelligent robots and systems}.\hskip 1em plus 0.5em minus 0.4em\relax IEEE, 2010, pp. 4584--4589.

\bibitem{claes2012collision}
D.~Claes, D.~Hennes, K.~Tuyls, and W.~Meeussen, ``Collision avoidance under bounded localization uncertainty,'' in \emph{2012 IEEE/RSJ International Conference on Intelligent Robots and Systems}.\hskip 1em plus 0.5em minus 0.4em\relax IEEE, 2012, pp. 1192--1198.

\bibitem{alonso2018cooperative}
J.~Alonso-Mora, P.~Beardsley, and R.~Siegwart, ``Cooperative collision avoidance for nonholonomic robots,'' \emph{IEEE Transactions on Robotics}, vol.~34, no.~2, pp. 404--420, 2018.

\bibitem{van2016online}
R.~Van~Parys and G.~Pipeleers, ``Online distributed motion planning for multi-vehicle systems,'' in \emph{2016 European Control Conference (ECC)}.\hskip 1em plus 0.5em minus 0.4em\relax IEEE, 2016, pp. 1580--1585.

\bibitem{luis2020online}
C.~E. Luis, M.~Vukosavljev, and A.~P. Schoellig, ``Online trajectory generation with distributed model predictive control for multi-robot motion planning,'' \emph{IEEE Robotics and Automation Letters}, vol.~5, no.~2, pp. 604--611, 2020.

\bibitem{ferranti2022distributed}
L.~Ferranti, L.~Lyons, R.~R. Negenborn, T.~Keviczky, and J.~Alonso-Mora, ``Distributed nonlinear trajectory optimization for multi-robot motion planning,'' \emph{IEEE Transactions on Control Systems Technology}, vol.~31, no.~2, pp. 809--824, 2022.

\bibitem{tordesillas2020mader}
J.~Tordesillas and J.~P. How, ``{MADER}: Trajectory planner in multi-agent and dynamic environments,'' \emph{IEEE Transactions on Robotics}, 2021.

\bibitem{li2020graph}
Q.~Li, F.~Gama, A.~Ribeiro, and A.~Prorok, ``Graph neural networks for decentralized multi-robot path planning,'' in \emph{2020 IEEE/RSJ international conference on intelligent robots and systems (IROS)}.\hskip 1em plus 0.5em minus 0.4em\relax IEEE, 2020, pp. 11\,785--11\,792.

\bibitem{niu2021multi}
Y.~Niu, R.~R. Paleja, and M.~C. Gombolay, ``Multi-agent graph-attention communication and teaming.'' in \emph{AAMAS}, vol.~21, 2021, p. 20th.

\bibitem{s23125615}
X.~Ye, Z.~Deng, Y.~Shi, and W.~Shen, ``Toward energy-efficient routing of multiple agvs with multi-agent reinforcement learning,'' \emph{Sensors}, vol.~23, no.~12, p. 5615, 2023.

\bibitem{10368056}
M.~Zhang and C.~Pan, ``Hierarchical optimization scheduling algorithm for logistics transport vehicles based on multi-agent reinforcement learning,'' \emph{IEEE Transactions on Intelligent Transportation Systems}, vol.~25, no.~3, pp. 3108--3117, 2024.

\bibitem{paul2023efficient}
S.~Paul, W.~Li, B.~Smyth, Y.~Chen, Y.~Gel, and S.~Chowdhury, ``Efficient planning of multi-robot collective transport using graph reinforcement learning with higher order topological abstraction,'' in \emph{2023 IEEE International Conference on Robotics and Automation (ICRA)}.\hskip 1em plus 0.5em minus 0.4em\relax IEEE, 2023, pp. 5779--5785.

\bibitem{sharma2023d2coplan}
V.~D. Sharma, L.~Zhou, and P.~Tokekar, ``D2coplan: A differentiable decentralized planner for multi-robot coverage,'' in \emph{2023 IEEE International Conference on Robotics and Automation (ICRA)}.\hskip 1em plus 0.5em minus 0.4em\relax IEEE, 2023, pp. 3425--3431.

\bibitem{chen2024scalable}
Y.~Chen, J.~Arkin, Y.~Zhang, N.~Roy, and C.~Fan, ``Scalable multi-robot collaboration with large language models: Centralized or decentralized systems?'' in \emph{2024 IEEE International Conference on Robotics and Automation (ICRA)}.\hskip 1em plus 0.5em minus 0.4em\relax IEEE, 2024, pp. 4311--4317.

\bibitem{RobotWeb}
R.~Murai, J.~Ortiz, S.~Saeedi, P.~H. Kelly, and A.~J. Davison, ``A robot web for distributed many-device localisation,'' \emph{IEEE Transactions on Robotics}, 2023.

\bibitem{dellaert_factor_2021}
F.~Dellaert, ``Factor graphs: Exploiting structure in robotics,'' \emph{Annual Review of Control, Robotics, and Autonomous Systems}, vol.~4, no.~1, pp. 141--166, 2021.

\bibitem{ortiz_visual_2021}
J.~Ortiz, T.~Evans, and A.~J. Davison, ``A visual introduction to gaussian belief propagation,'' \emph{arXiv preprint arXiv:2107.02308}, 2021.

\bibitem{karaman2011sampling}
S.~Karaman and E.~Frazzoli, ``Sampling-based algorithms for optimal motion planning,'' \emph{The international journal of robotics research}, vol.~30, no.~7, pp. 846--894, 2011.

\end{thebibliography}


\end{document}